\documentclass{article}
\PassOptionsToPackage{numbers, compress}{natbib}

\usepackage[preprint]{neurips_2025}

\usepackage[utf8]{inputenc}
\usepackage[T1]{fontenc}
\usepackage{hyperref}
\usepackage{url}
\usepackage{booktabs}
\usepackage{ragged2e}
\usepackage{amsfonts}
\usepackage{nicefrac}
\usepackage{microtype}
\usepackage{xcolor} 
\usepackage{svg}

\usepackage{array}
\usepackage{algorithm}
\usepackage{algpseudocode}

\usepackage{booktabs}   % for \toprule, \midrule, \bottomrule
\usepackage{multirow}   % for \multirow

\usepackage{amsmath}
\usepackage{markdown}

\title{DualSchool: \\ How Reliable are LLMs for Optimization Education?}

\newcommand{\aifouropt}{NSF AI Institute for Advances in Optimization}
\newcommand{\gatech}{Georgia Institute of Technology}
\newcommand{\atlanta}{Atlanta, GA, USA}

\newcommand*\samethanks[1][\value{footnote}]{\footnotemark[#1]}
\newcommand{\corresp}{\thanks{Corresponding author: \texttt{klam@isye.gatech.edu}}\,\;}
\newcommand{\equalcontrib}{\thanks{Equal contribution}\,\;}
\author{%
  Michael Klamkin\corresp\equalcontrib\thanks{\aifouropt, \gatech, \atlanta} \\
  \And
  Arnaud Deza$^\dagger$\samethanks \\
  \AND
  Sikai Cheng\samethanks \\
  \And
  Haoruo Zhao\samethanks \\
  \And 
  Pascal Van~Hentenryck\samethanks
}

\begin{document}

\maketitle
    
%%%%%%%%%%%%%%%%%%%%%%%%%%%%%%%%%%%%%%%%%%%%%%%%%%%%%%%%
%                       Main text
%%%%%%%%%%%%%%%%%%%%%%%%%%%%%%%%%%%%%%%%%%%%%%%%%%%%%%%%
\begin{abstract}
Consider the following task taught in introductory
optimization courses which addresses challenges articulated by the community at the intersection of (generative) AI and OR: {\em generate the dual of a linear program}. LLMs, being trained at web-scale, have the conversion process and many instances of Primal to Dual Conversion (P2DC) at their disposal. Students may thus reasonably expect that LLMs would perform well on the P2DC task. To assess this expectation, this paper introduces \textsc{{DualSchool}}, a comprehensive framework for generating and verifying P2DC instances. The verification procedure of \textsc{{DualSchool}} uses the {\em Canonical Graph Edit Distance}, going well beyond existing evaluation methods for optimization models, which exhibit many false positives and negatives when applied to P2DC. Experiments performed by \textsc{{DualSchool}} reveal interesting findings. Although LLMs can recite the conversion procedure accurately, state-of-the-art open LLMs fail to consistently produce correct duals. This finding holds even for the smallest two-variable instances and for derivative tasks, such as correctness, verification, and error classification. The paper also discusses the implications for educators, students, and the development of large reasoning systems.
\end{abstract}

\section{Introduction}\label{sec:intro}

Large Language Models (LLMs) have garnered significant interest for their potential to serve as always-available personalized education assistants, automating time-consuming tasks such as tutoring and grading in STEM education. To fully realize this potential, however, LLMs must demonstrate the ability to reliably execute detailed multi-step procedures. In particular, real-world tasks are often nuanced, making attention to detail paramount to success -- a higher bar than plausible-looking text.

This paper proposes the relatively simple primal-to-dual conversion (P2DC)  task as a benchmark to evaluate whether LLMs can execute detailed procedures reliably. P2DC is an interesting task for several reasons. (1) P2DC is commonly taught in introductory optimization courses.
(2) LLMs, being trained on a web scale, have the conversion process and many instances of Primal to Dual Conversion (P2DC) in their training corpus. Indeed, when asked directly how to do P2DC, most LLMs respond with a correct strategy. (3) P2DC requires a clear understanding of the procedure, since there are many ways to obtain a dual. (4) P2DC captures several challenges recently articulated at the intersection of (generative) AI and OR, including the verification of optimization models, the availability of datasets, and the design of evaluation criteria and methodologies \cite{SegevSlides}. (5) P2DC is also an inherently structured task since the input and output are linear programs which can be represented in different formats (e.g., JSON, XML, MPS files). This makes the P2DC task an attractive test-bed for reasoning models specialized to structured data, a relatively under-studied but extremely valuable competency. 

\begin{figure}[!t]
    \centering
    \includegraphics[width=\linewidth]{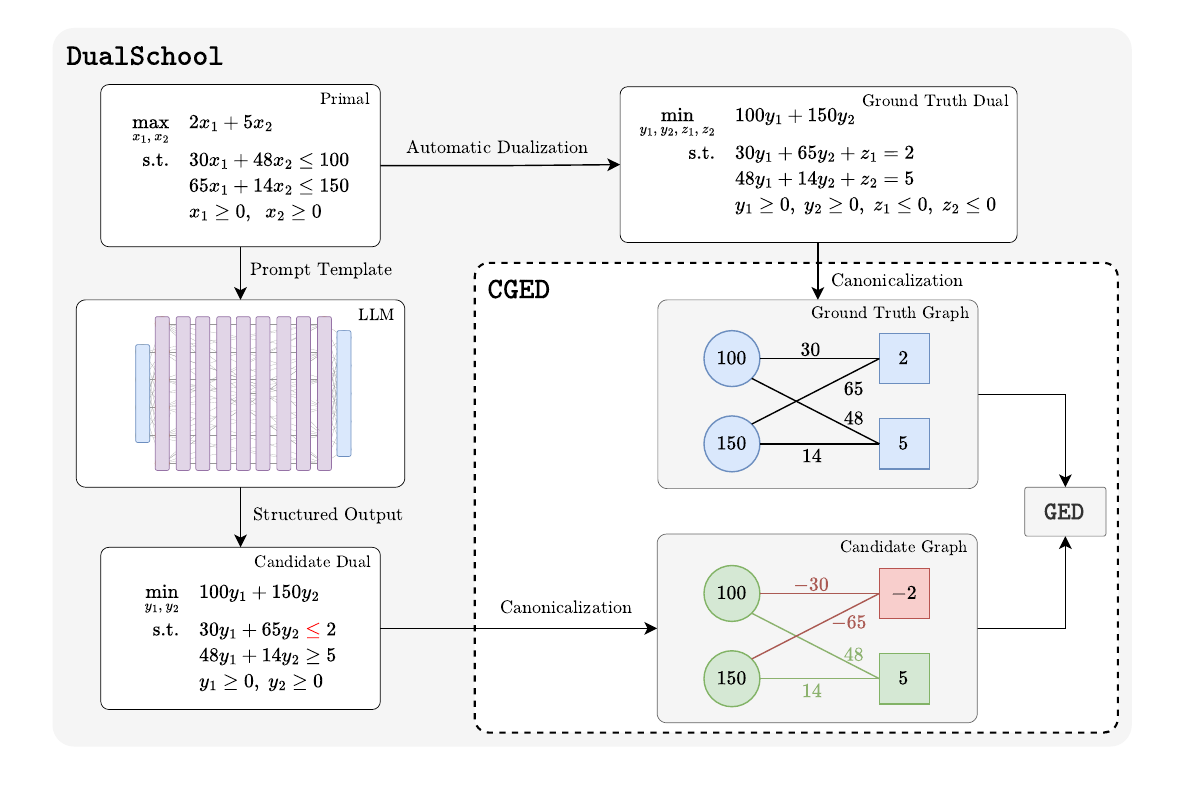}
    \caption{The P2DC task of {\sc DualSchool}: it illustrates the primal-to-dual conversion, the canonical representation of linear programs and the evaluation using CGED, which is the concatenation of the canonicalization step and the Graph Edit Distance comparison.}
    \label{fig:dualschool}
\end{figure}

Because of the simplicity of P2DC and the availability of the P2DC instructions and instances in the training corpus of LLMs, students in optimization classes may reasonably expect that LLMs would perform well on the P2DC task. To assess this expectation, this paper introduces \textsc{{DualSchool}}, a comprehensive framework for generating and verifying P2DC instances. To generate P2DC instances, {\sc DualSchool} leverages automatic symbolic dualization, converting new synthetic and existing primal-only datasets (e.g. NLP4OPT \cite{pmlr-v220-ramamonjison23a}, ComplexOR \cite{xiao2023chain}, and EasyLP \cite{huang2024mamo}), to P2DC datasets. To enable automatic evaluation, DualSchool includes a graph-based equivalence detection algorithm called {\em Canonical Graph Edit Distance (CGED)}. CGED is similar to the NGED algorithm of \citet{xing2024towards}, but it adds a crucial pre-processing canonicalization step specifically designed to allow for differences in dualization procedure conventions. As such, {\sc DualSchool} overcomes the limitations of existing validation techniques in the optimization setting which are either overly restrictive or overly permissive. Indeed, existing validations either force particular convention choices or forget much of the problem structure, complicating their use in post-training techniques to improve performance (e.g., \cite{Rao,jha2024rlsf}). P2DC is illustrated in Figure \ref{fig:dualschool}, which exemplifies the canonical form used for comparing linear programs.

Preliminary experimental results with \textsc{{DualSchool}} reveal interesting findings: they show that the P2DC is surprisingly challenging for leading open LLMs. This discrepancy -- being able to recite the procedure but not carry it out reliably -- underscores a critical limitation of LLMs: they yield duals with mistakes that may be minor in terms of token count but are clearly wrong (e.g., an unbounded dual for a feasible primal). These findings hold even for the smallest two-variable instances and for derivative tasks, such as error correction, error classification, and verification. In {\sc correction}, the LLM is asked to correct the error; in {\sc classification}, the LLM is asked what the error is; and in {\sc verification}, the LLM is asked if the primal-dual pair is valid.

The main contributions of this paper can be summarized as follows:
\begin{enumerate}
    \item The paper proposes {\sc DualSchool}, a comprehensive framework to evaluate the reliability of LLMs for a relatively simple optimization task, whose instructions are widely available. The multi-task framework includes the P2DC (primal to dual conversion) task and derivative tasks \textsc{correction}, \textsc{classification}, and \textsc{verification}.
    \item The paper designs a {\em robust automatic evaluation} using a {\em Canonical Graph Edit Distance (CGED)} algorithm that simultaneously allows for differences in dualization convention while robustly checking the correctness of all problem data.
    \item The paper is associated with a repository of {\em open data and code}: over 1,300 primal-dual pairs as well as error-injected variants for each are published alongside the paper, including the code to automatically generate more samples if needed.
    \item Experimental results show that P2DC is a compelling challenge despite its simplicity, as state-of-the-art open LLMs struggle even for very small instances. P2DC, and its derivative tasks, also address several challenges recently articulated at the intersection of (generative) AI and OR \cite{SegevSlides}.
\end{enumerate}

\noindent
The rest of the paper is organized as follows. Section 
\ref{sec:relwork} discusses related work. Section \ref{sec:P2DC} introduces the P2DC task and Section \ref{sec:equivalence} introduces the CGED algorithm. Then, Section \ref{sec:experiments} presents the experimental results and Section \ref{sec:conclusion} concludes the paper.

\section{Related Work}\label{sec:relwork}
This section reviews related work at the intersection of large language models and optimization.

\paragraph{LLMs for Optimization} Recent years have witnessed a surge of interest in leveraging large language models (LLMs) for various optimization tasks (LLM4OPT). Notable examples include natural language modeling \cite{ahmaditeshnizi2024optimus,xiao2023chain,pmlr-v220-ramamonjison23a}, where LLMs are given a natural language description of an optimization problem and are asked to formulate it, conversational interfaces to configure and customize existing models \cite{tsouros2023holy,lawless2024want}, algorithmic configuration for cutting plane subroutines in MILP solvers \cite{lawless2024llms}, explaining infeasibilities \cite{chen2024diagnosing}, and even solving optimization problems directly \cite{yang2023large}. These efforts highlight the growing recognition of LLMs as a versatile tool that can be applied beyond traditional natural language processing tasks and into the realm of mathematical optimization.

\paragraph{Evaluation Methods} Despite the increasing attention, existing work in this area primarily rely on the optimal objective value as the sole criterion for evaluating correctness of LLM-generated optimization formulations. This approach has inherent limitations, as it can silently ignore major errors in the formulation such as missing or incorrect constraints/variables if those errors happen to not affect the optimal value. The problem of equivalence detection between different optimization formulations remains largely under-explored. \cite{xing2024towards} shows that their normalized graph edit distance (NGED) more closely aligns with human assessments of correctness than token and optimal-value based evaluation. However, in the context of P2DC, the direct use of NGED results in many false negatives due to benign conventional differences, while optimal value yields many false positives.

\paragraph{LLM for Education and Tutoring} The potential of LLMs in education and tutoring has attracted significant interest with several recent works exploring their use in personalized learning and automated assessment. \cite{miller2024llm} investigates the use of LLMs in personalized tutoring systems. Benchmarks such as MathTutorBench \cite{macina2025mathtutorbench} have been developed to evaluate the capabilities of LLMs in educational settings, and case studies such as \cite{mok2025using} and \cite{hickey2024large} explore their potential in physics and power engineering education respectively.

\section{Primal-to-Dual Conversion}
\label{sec:P2DC}

This section introduces P2DC, the main task of {\sc DualSchool}. A linear program (LP) is a constrained optimization problem with linear objective and affine constraints, i.e., a problem that can be stated as
\begin{subequations}
\label{eq:primal}
\begin{align}
\min_{x\in\mathbb{R}^n}\quad
& c^\top x\\
\text{s.t.}\quad
& a_j^\top x \le b_j&\quad\forall j\in \mathcal{I}_\le \label{eq:lpleq}\\
& a_j^\top x \ge b_j&\quad\forall j\in \mathcal{I}_\ge \label{eq:lpgeq}\\
& a_j^\top x = b_j&\quad\forall j\in \mathcal{I}_=\label{eq:lpeq}
\end{align}
\end{subequations}
where $x\in\mathbb{R}^n$ is the vector of decision variables, $c\in\mathbb{R}^n$ is the objective vector, $a_j\in\mathbb{R}^{n}$ are the constraint coefficient vectors, and $b_j\in\mathbb{R}$ are the constraint right-hand-sides. Any $x$ which satisfies \eqref{eq:lpleq}, \eqref{eq:lpgeq} and \eqref{eq:lpeq} (i.e., a \textit{feasible} $x$) provides an upper bound on the optimal value of the LP. Any feasible solution to the \textit{dual} of an LP, which itself is an LP\footnote{The dual program corresponding to Model \ref{eq:primal} is stated in Appendix \ref{appendix:p2dmethods} as Model \ref{eq:dual}.}, provides a lower bound on the optimal LP value. Moreover, in many practical applications, dual programs often have useful interpretations, as demonstrated in the example below. Readers are referred to \citet[Section 1.3]{nemirovski2024introduction} for a detailed introduction to linear programming duality. 

\paragraph{Example: Production Planning}
Consider the production planning problem where a factory manager is tasked with finding the most profitable production plan given a fixed amount of resources; wood ($W$) and steel ($S$). In this example, the factory can produce a number of doors ($d$) and tables ($t$), each using varying amounts of resources. All products are sold for a profit, $p_d$ for doors and $p_t$ for tables. The amount of wood (resp. steel) needed to produce a door is denoted by $a_{wd}$ (resp. $a_{sd}$) and likewise the amount of wood (resp. steel) needed to produce a table is denoted by $a_{wt}$ (resp. $a_{st}$). The full formulation is stated below as Model \ref{model:ppprimal}.
The dual of this program, given below as Model 
\ref{model:ppdual}, is known as the resource-valuation problem \cite{carsberg1969linear}, with variables $y_w$ and $y_s$ denoting the so-called ``shadow-price'' of wood and steel respectively.

\noindent
\begin{minipage}{.40\textwidth}
\begin{equation}\label{model:ppprimal}
\begin{aligned}
    \max_{d,t} \quad
    & p_d d + p_t t \\
    \text{s.t.} \quad
    & a_{wd}d + a_{wt}t \leq W \\
    & a_{sd}d + a_{st}t \leq S \\
    & d\geq0,\;\;t\geq0
\end{aligned}
\end{equation}
\end{minipage}
\begin{minipage}{.59\textwidth}
\begin{equation}
\implies\quad\quad\quad
\begin{aligned}
\label{model:ppdual}
    \min_{y_w,y_s} \quad
    & Wy_w + Sy_s \\
    \text{s.t.} \quad
    & a_{wd}y_w + a_{sd}y_s \geq p_d \\
    & a_{wt}y_w + a_{st}y_s \geq p_t \\
    & y_w\geq0,\;\;y_s\geq0
\end{aligned}
\end{equation}

\end{minipage}

\noindent
Performing this transformation -- converting Model \ref{model:ppprimal} to Model \ref{model:ppdual} -- is the main task in {\sc DualSchool}, called Primal-to-Dual Conversion (P2DC). Note that there exists several different procedures for deriving the dual of an LP; the most common methods are summarized in Appendix \ref{appendix:p2dmethods}.
The dual program is not unique in that there exist different procedures and convention choices that can yield different, but valid, dual programs. For instance, consider Model \ref{model:ppdual_slacked}, that differs from Model \ref{model:ppdual} only in its use of slack variables $z_d$ and $z_t$ to convert the inequality constraints to equalities. In the context of P2DC, one may arrive at Model \ref{model:ppdual_slacked} if dualizing by first converting the primal to the standard form $\min\;c^\top x\;\,\text{s.t.}\;\;Ax\leq b$ then writing the dual as $\max\; b^\top y\;\,\text{s.t.}\;\;A^\top y =c,\;y\geq0$.
\begin{equation}
\begin{aligned}
\label{model:ppdual_slacked}
    \min_{y_w,y_s, z_d, z_t} \quad
    & Wy_w + Sy_s \\
    \text{s.t.} \quad
    & a_{wd}y_w + a_{sd}y_s - z_d = p_d \\
    & a_{wt}y_w + a_{st}y_s - z_t = p_t \\
    & y_w\geq0,\;\;y_s\geq0,\;\;z_d\geq0,\;\;z_t\geq0
\end{aligned}
\end{equation}
Models \ref{model:ppdual} and \ref{model:ppdual_slacked} are both considered ``correct'' duals of Model \ref{model:ppprimal}. This complicates equivalence detection, since directly applying NGED on the graphs corresponding to Models \ref{model:ppdual} and \ref{model:ppdual_slacked} would result in a non-zero edit-distance due to the missing slack variable nodes, missing edges denoting the slack constraint coefficients, and changed constraint senses. Thus, in the context of P2DC, it is important to use a ``convention-invariant'' matching procedure that explicitly treats such differences. Section \ref{sec:equivalence} expands on the shortcomings of existing approaches and proposes a new metric, {\em Canonical Graph Edit Distance (CGED)}, that meets the requirements of the P2DC setting.

\section{Automatic Evaluation for P2DC}\label{sec:equivalence}
This section begins by explaining why existing evaluation methods are insufficient to determine the correctness of a candidate dual in the P2DC setting. It then proposes a correctness detection algorithm that extends that of \citet{xing2024towards} in order to enable a ``convention-invariant'' matching of the candidate dual program to a known correct dual formulation obtained by automatic dualization.

\paragraph{NER-based matching}
Several prior works \cite{pmlr-v220-ramamonjison23a,ramamonjison2022augmenting,prasath2023synthesis}, including the NL4OPT ``generation'' sub-task, use a declaration-level accuracy \cite[Equation 2]{pmlr-v220-ramamonjison23a} that matches tokens within declarations. Although, compared to a naive token-matching, this metric allows to handle permutations of declarations, it cannot handle basic symmetries that are either \textit{within} declarations, such as the order of terms in a constraint, or span across multiple declarations, such as variable sign convention.

\paragraph{Optimal Value} Most prior work uses an optimal value check, often referred to as execution accuracy, to establish the correctness of a given formulation \cite{ahmaditeshnizi2024optimus,huang2024orlm}. Although this is a necessary condition for correctness, it often yields false positives. For example, if the formulation omits a constraint that is not tight at the optimal solution, the optimal value check will still mark the formulation correct. Furthermore, in the P2DC setting, simply echoing back the primal model always gives the same objective value, even for problems which are not self-dual (due to strong duality). This makes the optimal value check easy to ``reward-hack'' \cite{rewardhack}, limiting its applicability as a reward signal.

\paragraph{Polyhedral congruence and isomorphism}
Since the feasible set of an LP is a polyhedron, polyhedral computation libraries, e.g., \texttt{polymake} \cite{polymake}, can be used to evaluate whether the polyhedra corresponding to the candidate and ground truth feasible sets are congruent or isomorphic. However, checking congruency does not fit the P2DC setting since, although it does treat permutations and variable sign convention differences and, in some cases it may be useful to forget constraint scaling and redundant constraints, polyhedral congruence also allows for arbitrary transformations such as rotation and translation, breaking the primal-dual correspondence. Similarly, polyhedral isomorphism is not a good fit since it verifies only the incidence structure of the polyhedron, effectively forgetting most of the structure imposed by the problem data.

\paragraph{Graph Edit Distance} Recent work in ML for optimization use graph representations of optimization problems \cite{gasse2019exact,gupta2020hybrid}. \citet{xing2024towards} proposes to use graph edit distance (GED) algorithms on these graph representations to evaluate equivalence between a candidate and a ground truth program. Their method, called NGED, is attractive due to its ability to handle variable and constraint permutations. Furthermore, GED is a rich reward signal in that it outputs not only a boolean value but the optimal edit path between the two formulations. However, NGED is still overly restrictive in the P2DC setting since dualization procedures and conventions can result in dual programs that have different graph representations, i.e. additional variable nodes and edges when slack variables appear. Thus, directly applying NGED is too restrictive for P2DC.

Finally, note that each of these metrics rely on successful parsing of the LLM output into the required representation for comparison, e.g. XML for NER or the bipartite graph for GED. In cases where the parser fails due to incorrect formatting of the LLM response, the paper deems the formulations not equivalent. However, as shown in Section \ref{sec:experiments}, the vast majority of responses are correctly parsed.

\subsection{Canonical Graph Edit Distance}

This paper proposes the {\em Canonical Graph Edit Distance (CGED)} for correctness detection: it modifies the NGED method to include a canonicalization step in order to 
control for variations in  dualization procedure. In particular, the paper notes that these convention differences result in (combinations of) two kinds of ``symmetries'' in the dual programs:
variable sign and slack variables. Although NGED itself includes some canonicalization such as converting the objective sense to minimization and single-sided inequalities to less-than sense, it fails to treat these convention differences, leading to many false negatives as demonstrated in Section \ref{sec:experiments}. The following paragraphs describe the proposed modifications in detail.

\paragraph{Slack variables}
As demonstrated using Models \ref{model:ppdual} and \ref{model:ppdual_slacked} in Section \ref{sec:P2DC}, slack variables appear in the dual when the dualization procedure treats primal variables as free, either by explicitly including their bounds in the main constraints or due to the particular standard form or ruleset used.
The correction detection algorithm treats these convention differences by eliminating the resulting slack variables. Equation \eqref{eq:slackelim} demonstrates such an elimination.
\begin{equation}\label{eq:slackelim}
    \begin{aligned}
    \min\quad& x_1+x_2 \\
    \text{s.t.}\quad&    x_1+x_2+s=1\\
        &s\leq 0
    \end{aligned}
    \quad\implies\quad
    \begin{aligned}
    \min\quad& x_1+x_2 \\
    \text{s.t.}\quad&    x_1+x_2\geq1
    \end{aligned}
\end{equation}

\paragraph{Variable Sign}
The second common symmetry that arises due to variations in dualization convention is variable sign, since as pointed out in Step 2 of the Lagrangian dualization method in Appendix \ref{appendix:p2dmethods}, the sign constraint given to a dual variable only has to match how the corresponding constraint's residual is formed. In other words, as long as the memorized procedure/standard form/ruleset is consistent, the dual variables can be given any sign, including  free variables (corresponding to equality constraints in the primal). For instance, the programs in Equation \ref{eq:signelim} are deemed equivalent for the purposes of P2DC.
\begin{equation}\label{eq:signelim}
    \begin{aligned}
    \min\quad& x_1 + x_2 \\
    \text{s.t.}\quad& x_1+x_2\geq1 \\    & x_1\geq0,\;x_2\geq0
    \end{aligned}
    \quad\iff\quad
    \begin{aligned}
    \min\quad& {-}x_1^\prime + x_2 \\
    \text{s.t.}\quad&    {-}x_1+x_2\geq1\\
 &x_1^\prime\leq0,\;x_2\geq0
    \end{aligned}
\end{equation}
To establish equivalence up to variable sign, both the candidate and ground truth are reformulated to convert variables whose bound constraint reads $x\leq u$ to $x\geq -u$ by flipping the sign of its constraint and objective coefficients. In order to allow for sign convention differences in free variables and those with double-sided finite bounds, CGED exploits the variable permutation property of GED by using the difference-of-positive variables transformation $x\in\mathbb{R}\implies x = x^+-x^-,\;x^+\geq0,\,x^-\geq0$. An example for the double-sided finite bounded variable case is given in Appendix \ref{appendix:dsbc}.

When these modifications are used as pre-processing steps for GED, the overall procedure is able to recognize the correctness of a dual even if it differs from the ground truth in the following ways:
\begin{enumerate}
    \item \textbf{Objective sense} -- flipping the objective sense is allowed as long as all objective coefficient signs are also flipped. In P2DC, this corresponds to a common post-processing that is applied if, for instance, all the objective coefficients are negative.
    \item \textbf{Inequality sense} -- flipping the sense of an inequality is allowed as long as the signs of the coefficients and RHS are also flipped. In P2DC this corresponds to a common post-processing that is applied if, for instance, all the problem data in a constraint is negative.
    \item \textbf{Variable and constraint permutation} -- reordering constraints and variables is allowed.
    \item \textbf{Slack variables} -- using slack variables to turn inequalities into equalities is allowed. In P2DC, slack variables appear in the dual when treating primal variables as free.
    \item \textbf{Variable sign} -- flipping the sign of a variable is allowed as long as the sign of its constraint and objective coefficients are also flipped. In P2DC, this corresponds to a common convention choice when defining dual variables associated with primal inequality constraints.
\end{enumerate}

Note that although CGED is designed specifically for the P2DC setting, the canonicalization procedures can be used more broadly to detect equivalence between formulations or as a normalization procedure for systems that take a linear program as input. For other applications, it is important to consider exactly what should be preserved and what should be forgotten. In particular, due to its specialization to P2DC, CGED does not treat several symmetries which may be natural to forget in other areas such as scaling of variables or constraints and variable substitutions.

\section{Experiments}\label{sec:experiments}
This section describes the experiments conducted to evaluate the performance of leading open LLMs on the {\sc DualSchool} dataset. 

\paragraph{Benchmark Instances}
{\sc DualSchool} comprises over 1300 LP instances drawn from three main sources: (1) two dimensional LPs from bounded toy poltyopes, (2) continuous relaxations of small-scale combinatorial optimization instances and (3) LP instances from prior work on natural language modeling benchmarks\cite{pmlr-v220-ramamonjison23a,xiao2023chain,huang2024mamo,ahmaditeshnizi2024optimus}. 
For each instance, {\sc DualSchool} includes ground-truth duals generated using symbolic dualization \cite{dualizationjl}. For the \textsc{Correction}, \textsc{Verification}, and \textsc{Classification} tasks, {\sc DualSchool} also includes duals with (labeled) errors; the error types are described in Appendix \ref{appendix:P2D_errors}. Table~\ref{tab:datasets_sizes} summarizes the data sources.

\begin{table}[t]
\centering
\caption{Mean (max) number of variables and constraints for each dataset.}
\label{tab:datasets_sizes}
\resizebox{\textwidth}{!}{%
  \begin{tabular}{l c c c c c c}
  \toprule
   & \textbf{ComplexOR}\cite{xiao2023chain} & \textbf{Easy LP} \cite{huang2024mamo} & \textbf{NL4OPT}\cite{pmlr-v220-ramamonjison23a} & \textbf{NLP4LP}\cite{ahmaditeshnizi2024optimus} & \textbf{2D} & \textbf{CO-Small} \\
  \midrule
  \textbf{Variables}   & 4.1 (9)  & 2.8 (5) & 2.0 (3) & 2.2 (6) & 2.0 (2) & 3.9 (5)  \\
  \textbf{Constraints} & 4.5 (12) & 4.3 (14)& 2.9 (5)& 3.1 (6)& 5.7 (12) & 3.5 (6)  \\
  \hline\noalign{\smallskip}
  \textbf{Instances}   & 15    & 585 & 205 & 266 & 108 & 140   \\
  \bottomrule
  \end{tabular}%
}
\end{table}

\paragraph{Language Models} 
Due to resource limitations, the experiments consider only small and medium-sized open-weight LLMs.
The models are evaluated in both the zero‐shot and one‐shot in‐context learning settings. Readers are referred to Appendix \ref{appendix:exp_setup_cont} for more details about model configurations and compute resources.

\paragraph{Evaluation Pipeline} For \textsc{Generation} and \textsc{Correction}, the LLM is prompted to produce \texttt{gurobipy} code that formulates the dual. The code is executed in a sandbox environment to create the \texttt{gurobipy.Model} object which is then written to MPS for evaluation. That MPS file is then compared to that of the ground truth dual using NGED \cite{xing2024towards}, OBJ \cite{ahmaditeshnizi2024optimus,huang2024orlm}, and the proposed CGED (Section \ref{sec:equivalence}).  
Instances that crash or cannot be parsed are counted as incorrect. For the {\sc verification} and {\sc classification} tasks, two methods for extracting an answer from the LLM response are tested: 1) XML from free-flow output and 2) enforced JSON schema.\footnote{using the Structured Outputs feature in Ollama \url{https://ollama.com/blog/structured-outputs}}
Problems are rendered to the model in \LaTeX \ using the prompt template described in Appendix \ref{appendix:prompt}.\footnote{\LaTeX \ is generated using \url{https://jump.dev/JuMP.jl/stable/manual/models/\#Print-the-model}}

 The {\sc DualSchool} tasks have two kinds of outputs: models (\textsc{Generation} and \textsc{Correction}) and choices (\textsc{Verification} and \textsc{Correction}). For tasks with model outputs, three metrics are reported: the canonical graph edit distance (CGED) from Section \ref{sec:equivalence}, the normalized graph edit distance (NGED) from \citet{xing2024towards}, and  the objective-match (OBJ). Each of these is reported as an accuracy, i.e. how often  the edit distance to the ground truth dual is equal to zero. For tasks with choice outputs, the classification accuracy is reported, i.e. how often the LLM chose the correct choice.
 
\subsection{Results for P2DC \textsc{Generation}}

Table~\ref{tab:llm-benchmark} reports the CGED, NGED, and OBJ accuracies across four benchmark datasets under both 0-shot and 1-shot prompting conditions. The Execution accuracy (Exec\%) column reports the percentage of instances for which the LLM code successfully produced an MPS file. Overall, even though Exec\% is high for most models, no model reliably produces correct duals -- even when prompted with small, synthetic instances. The best-performing model, Phi 4 - 14B, reaches 47.8\% CGED and  53.7\% objective accuracy on the NL4OPT samples in the 0-shot setting. Due to space limitations and relatively poor performance, the results for the instances coming from CO small and Easy LP are presented in Appendix \ref{appendix:exp_setup_cont} in Table \ref{tab:llm-benchmark_additional}. 

In all cases, OBJ consistently exceeds CGED. This highlights a key pitfall of objective-based evaluation: LLMs often produce duals with the correct objective value but incorrect or malformed structure. Based on an informal analysis of samples in this category -- CGED is nonzero while OBJ is true -- the most common mistake is omitting a (dual) variable bound that happened to be redundant. Conversely, NGED is too restrictive in this setting, giving consistently lower scores than CGED. These results underscore the need for convention-invariant evaluation like CGED in the P2DC setting. Surprisingly, one-shot prompting provides no consistent benefit and occasionally degrades performance (e.g.\ Phi-4 drops from 35.7\% to 28.6\% on the \textsc{ComplexOR} samples), suggesting limited applicability of in-context learning in the P2DC setting.

\begin{table}[t]
  \centering
  \caption{Aggregated accuracy results for the {\sc Generation} task}
  \label{tab:llm-benchmark}
  \resizebox{\textwidth}{!}{%
    \begin{tabular}{llcc@{\hspace{0.75\tabcolsep}}c@{\hspace{0.75\tabcolsep}}cc@{\hspace{0.75\tabcolsep}}c@{\hspace{0.75\tabcolsep}}cc@{\hspace{0.75\tabcolsep}}c@{\hspace{0.75\tabcolsep}}cc@{\hspace{0.75\tabcolsep}}c@{\hspace{0.75\tabcolsep}}c}
      \toprule
       &  &  & \multicolumn{3}{c}{\textbf{ComplexOR}} & \multicolumn{3}{c}{\textbf{NL4OPT}} & \multicolumn{3}{c}{\textbf{NLP4LP}} & \multicolumn{3}{c}{\textbf{2D}} \\
      \cmidrule(lr){4-6} \cmidrule(lr){7-9} \cmidrule(lr){10-12} \cmidrule(lr){13-15}
      \textbf{Model} & \textbf{Prompt} & \textbf{Exec\%} & NGED & OBJ & CGED & NGED & OBJ & CGED & NGED & OBJ & CGED & NGED & OBJ & CGED \\
      \midrule
      \multirow{2}{*}{Mistral-7B} & 0-shot & 24.8 & 0 & 0 & 0 & 0 & 0 & 0 & 0 & 3.9 & 0 & 0 & 11.5 & 0 \\
      & 1-shot & 22.5 & 0 & 50.0 & 0 & 0 & 0 & 0 & 0 & 3.3 & 0 & 0 & 7.7 & 0 \\
      \midrule
      \multirow{2}{*}{Phi 4-14B} & 0-shot & 99.1 & 7.1 & 50.0 & \textbf{35.7} & 24.4 & 53.7 & \textbf{47.8} & 10.5 & 46.8 & \textbf{30.8} & 1.0 & 5.7 & \textbf{1.0} \\
      & 1-shot & 99.7 & 7.1 & 42.9 & \textbf{28.6} & 2.0 & 34.5 & \textbf{27.6} & 2.9 & 34.0 & \textbf{18.9} & 14.0 & 18.7 & \textbf{14.0} \\
      \midrule
      \multirow{2}{*}{Gemma 3-12B} & 0-shot & 69.9 & 0 & 0 & 0 & 0.6 & 3.7 & 3.0 & 0 & 4.6 & 0 & 0 & 8.5 & 0 \\
      & 1-shot & 92.5 & 0 & 0 & 0 & 0 & 0 & 0 & 0 & 3.0 & 0 & 0 & 5.8 & 0 \\
      \midrule
      \multirow{2}{*}{Qwen 2.5–7B} & 0-shot & 93.7 & 0 & 0 & 0 & 0 & 3.1 & 0.5 & 1.3 & 3.5 & 1.8 & 0 & 1.9 & 0 \\
      & 1-shot & 94.2 & 0 & 0 & 0 & 0 & 0 & 0 & 0 & 1.8 & 0 & 0 & 2.9 & 0 \\
      \midrule
      \multirow{2}{*}{Qwen 2.5–14B} & 0-shot & 92.2 & 7.1 & 14.3 & 7.1 & 4.3 & 20.7 & 13.6 & 2.7 & 10.9 & 3.6 & 0 & 2.1 & 0 \\
      & 1-shot & 93.6 & 0 & 0 & 0 & 0 & 0 & 0 & 0 & 3.1 & 0.4 & 10.4 & 23.6 &  10.4 \\
      \midrule
      \multirow{2}{*}{Llama 3.1-8B} & 0-shot & 94.8 & 0 & 0 & 0 & 0 & 1.7 & 1.1 & 0 & 1.4 & 0 & 0 & 6.6 & 0 \\
      & 1-shot & 95.3 & 0 & 0 & 0 & 0 & 0 & 0 & 0 & 4.6 & 1.4 & 0 & 9.7 & 0 \\
      \midrule
      \multirow{2}{*}{Llama 3.3-70B} & 0-shot & 73.8 & 16.7 & 41.7 & 25.0 & 17.0 & 45.8 & 39.9 & 8.2 & 30.6 & 18.9 & 0 & 11.1 & 0 \\
      & 1-shot & 100.0 & 0 & 21.4 & 14.3 & 0 & 19.0 & 17.6 & 0.4 & 24.8 & 16.0 & 1.9 & 6.5 & 1.9 \\
      \midrule
      \bottomrule
    \end{tabular}
  }
\end{table}

\subsection{Results for P2DC \textsc{Correction}}  
\begin{figure}[ht]
     \centering
     \includegraphics[width=0.86\textwidth]{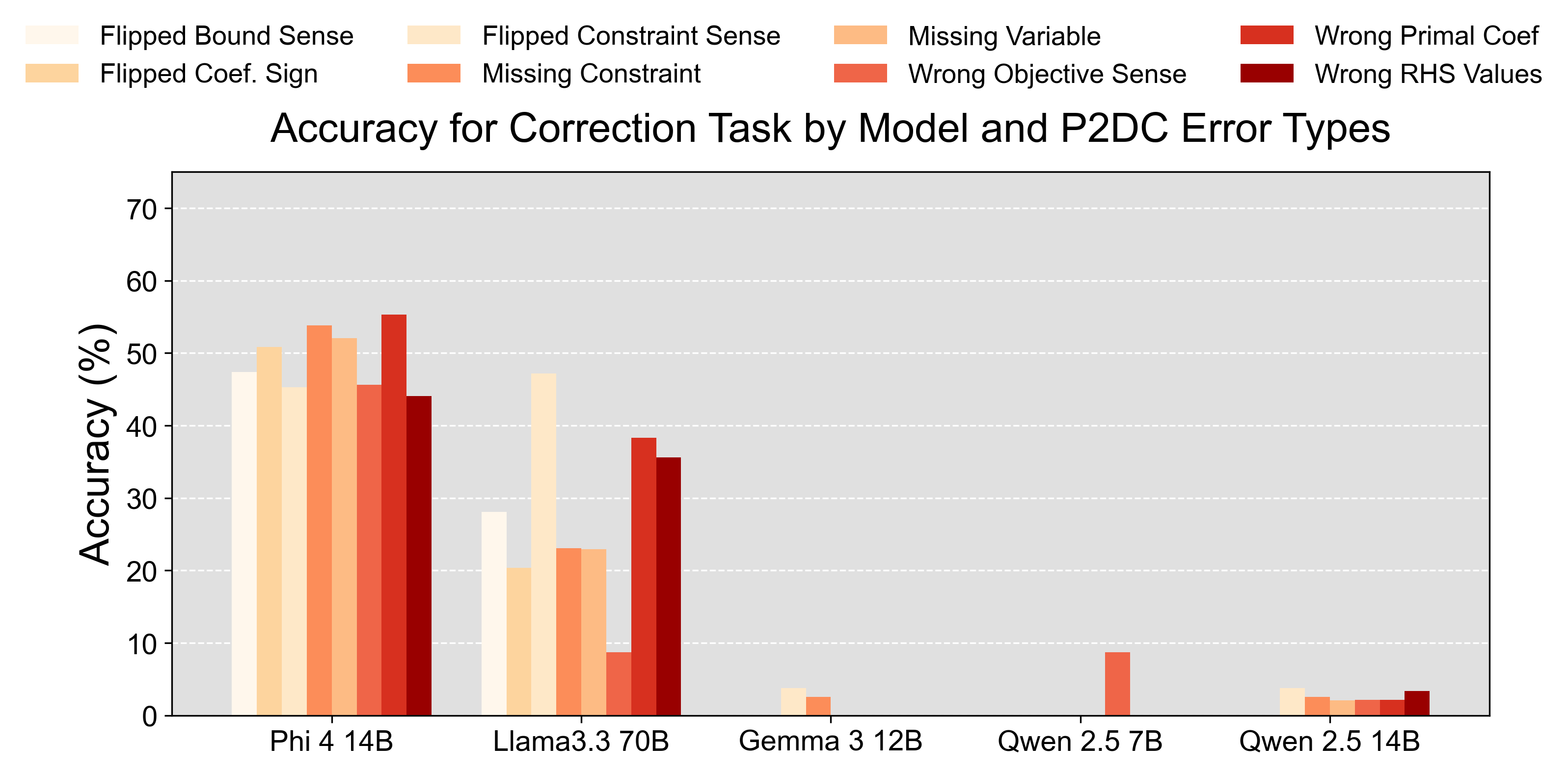}
     \caption{Accuracy for the \textsc{Correction} task by model and error type.}
     \label{fig:correction_acc}
\end{figure}

Figure \ref{fig:correction_acc} reports the performance of LLMs on the {\sc correction} task. All models struggle to reliably repair the incorrect duals, with accuracies below 60\% across all models and error types. Similarly to the {\sc generation} task, the Phi 4 and Llama 3.3 models outperform the others.
These uniformly low accuracies -- even on error types that are relatively easy to detect as shown in the next section -- reveal that {\sc correction} is essentially just as challenging as {\sc generation} for the open LLMs considered.
 
\subsection{Results for P2DC \textsc{Verification} and \textsc{Classification}} \label{sec:verification_classification}
 
\begin{figure}[t]
     \centering
     \includegraphics[width=\textwidth]{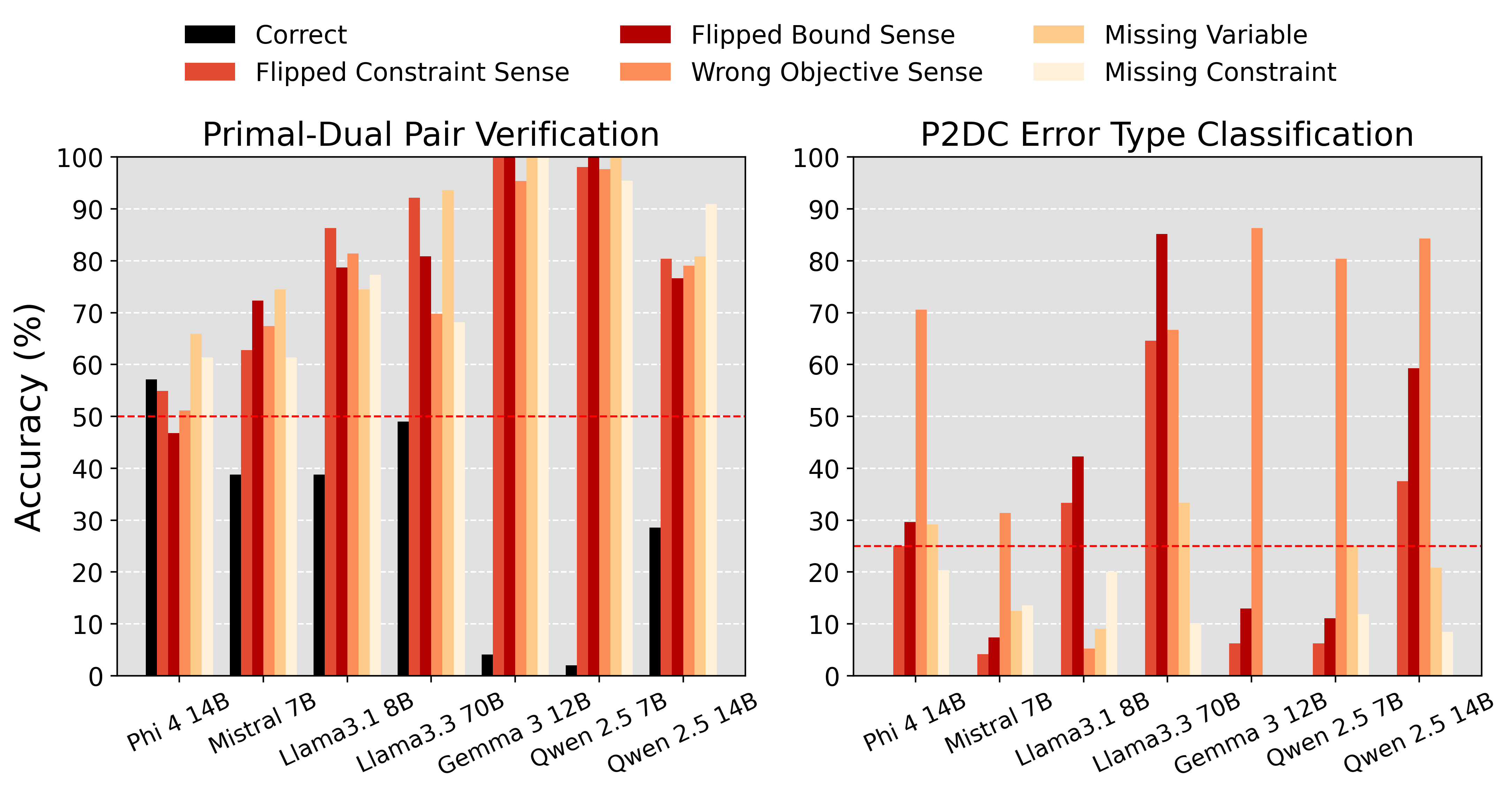}
     \caption{Accuracy for the \textsc{Verification} and \textsc{Classification} tasks by model and error type.}
     \label{fig:verification_acc}
\end{figure}

Figure~\ref{fig:verification_acc} reports the accuracies for the primal–dual pair {\sc verification} task (left) and the {\sc classification} task (right), with a red dashed line denoting the random‐guess baseline (50\% for {\sc verification}, 25\% for {\sc classification}). Overall, results are slightly negative: most models cluster at or below the random-guess baseline, reflecting limited reliability of predictions. Importantly, note that in the \textsc{Verification} task, across all models there is a clear bias towards predicting ``no'', resulting in high accuracy on all but the ``Correct'' category (black); the only one where the right answer is ``yes''. This is even more prevalent when using structured outputs, as shown in Appendix \ref{sec:more_exp} in Figure \ref{fig:verification_acc_structured}.

In the error classification task, detecting the flipped objective sense stands out as the easiest error type, with the best-performing models achieving accuracies between 70\% and 90\%. However, the more nuanced error types remain challenging with most models' accuracies hovering near or below the random‐guess line.

\section{Conclusion}\label{sec:conclusion} 
This paper introduced \textsc{{DualSchool}}, the first comprehensive benchmark for probing an LLM’s ability to perform and critique primal-to-dual conversions in linear programming. {\sc DualSchool} combines four structured tasks (generation, verification, correction, and error classification) with a graph-based correctness detector that goes beyond simple token matching or objective-value checks, and is specifically designed to avoid false positives and negatives for the primal-to-dual task. Code is published alongside the paper as well as a dataset of P2DC samples based on both synthetic and real-world LPs, each including a set of duals with injected errors. Preliminary experiments using leading open models show that the best LLMs tested achieve a best-case dualization accuracy of only $47.8\%$, with similarly low performance on the derivative tasks.

From an education standpoint, it is important to communicate to students, especially those without knowledge of generative AI and its implementation, that LLMs are not in the same equivalence class as e.g. Matlab or Julia. Although LLMs can return the "recipe" for various mathematical tasks, they struggle to follow these recipes even for simple tasks that are expected from undergraduate students in introductory classes. Moreover, it is important to be aware of the fact that LLM responses often feature high-quality writing and rich formatting, making it easy to believe the response is accurate.

From a research standpoint, {\sc DualSchool} provides a simple, yet meaningful benchmark to measure progress in LLMs and reasoning systems over the next years. Furthermore, thanks to the automatic labeling and evaluation algorithms, {\sc DualSchool} can be directly used in fine-tuning methods such as reinforcement learning with symbolic feedback \cite{jha2024rlsf} that can leverage rich reward signals. Future directions include extending {\sc DualSchool} to quadratic and conic formulations and evaluating its efficacy as a fine-tuning dataset.

%%%%%%%%%%%%%%%%%%%%%%%%%%%%%%%%%%%%%%%%%%%%%%%%%%%%%%%%
%                   Acknowledgements
%%%%%%%%%%%%%%%%%%%%%%%%%%%%%%%%%%%%%%%%%%%%%%%%%%%%%%%%
\newcommand{\acknowledgements}{\section*{Acknowledgements}This research was partly funded by NSF awards 2112533 and DGE-2039655. Any opinions, findings, and conclusions or recommendations expressed in this material are those of the author(s) and do not necessarily reflect the views of NSF. Experiments were run on the PACE Phoenix \cite{PACE} cluster.}
\makeatletter
\@ifpackagewith{neurips_2025}{preprint}{\acknowledgements}{}
\@ifpackagewith{neurips_2025}{final}{\acknowledgements}{}
\makeatother

%%%%%%%%%%%%%%%%%%%%%%%%%%%%%%%%%%%%%%%%%%%%%%%%%%%%%%%%
%                     Bibliography
%%%%%%%%%%%%%%%%%%%%%%%%%%%%%%%%%%%%%%%%%%%%%%%%%%%%%%%%
\bibliography{bibliography}
\bibliographystyle{unsrtnat}

%%%%%%%%%%%%%%%%%%%%%%%%%%%%%%%%%%%%%%%%%%%%%%%%%%%%%%%%
%                       Appendix
%%%%%%%%%%%%%%%%%%%%%%%%%%%%%%%%%%%%%%%%%%%%%%%%%%%%%%%%
\appendix
\newpage
\section{Appendix}

\subsection{Primal to dual conversion methods}\label{appendix:p2dmethods}

\paragraph{Standard Form}
A common method for forming the dual of a primal program is to first memorize a standard-form primal-dual pair, i.e. $\min\;c^\top x\;\,\text{s.t.}\;Ax\leq b \implies \max\;b^\top y\;\,\text{s.t.}\;A^\top y = c,\; y \geq 0$, then convert the given primal to that standard form and apply the memorized map. If the standard form primal memorized is Model \ref{eq:primal}, this method is coincides with SOB (modulo objective sense). The dual of Model \ref{eq:primal} is included below as Model \ref{eq:dual}.
\begin{subequations}
\label{eq:dual}
\begin{align}
\max_{x\in\mathbb{R}^n}\quad
& \sum_{j\in\mathcal{I}_\leq}b_j y_j + \sum_{j\in\mathcal{I}_lgeq}b_j y_j + \sum_{j\in\mathcal{I}_=}b_j y_j\\
\text{s.t.}\quad
& A^\top y = c \\
& y_j \geq 0 & \quad \forall j \in \mathcal{I}_\leq\\
& y_j \leq 0 & \quad \forall j \in \mathcal{I}_\geq\\
& y_j \in \mathbb{R} & \quad \forall j \in \mathcal{I}_=
\end{align}
\end{subequations}

\paragraph{Sensible-Odd-Bizarre}

\citet{benjamin1995sensible} describes the Sensible-Odd-Bizarre (SOB) method for remembering how to write the dual of a linear program. Variants of the method, i.e. table or rule-based approaches, are widely used as a practical approach to write the dual without having to go through a standard form.
The method is summarized in Table \ref{tab:sob}
\setlength{\extrarowheight}{0.25\baselineskip}
\begin{table}[h!]\label{tab:sob}
\centering
\caption{The Sensible-Odd-Bizarre method for mnemonic dualization \cite{benjamin1995sensible}.}
\begin{tabular}{|c|>{\centering\arraybackslash}p{5cm}|>{\centering\arraybackslash}p{5cm}|}
\hline

  & \textbf{Primal (Dual)}
  & \textbf{Dual (Primal)} \\
\hline
Objective
    & Maximize $c^\top x$ (Minimize $b^\top y$)
    & Maximize $b^\top y$ (Minimize $c^\top x$) \\
\hline
    & \RaggedRight Constraint $j$:
    & \RaggedRight Variable $y_j$ (or $x_i$): \\ 
Sensible
    & $a_j^\top x_i\geq b_j$
    & $y_j \geq 0$ \\
Odd
    & $a_j^\top x_i= b_j$
    & $y_j\in\mathbb{R}$ \\
Bizarre
    & $a_j^\top x_i\leq b_j$
    & $y_j \leq 0$ \\
\hline
    & \RaggedRight Variable $x_i$ (or $y_j$):
    & \RaggedRight Constraint $j$: \\ 
Sensible
    & $x_i \ge 0$
    & $a_i y_j\leq b_i$ \\
Odd
    & $x_i\in\mathbb{R}$
    & $a_i y_j= b_i$ \\
Bizarre
    & $x_i \le 0$
    & $a_i y_j\geq b_i$ \\
\hline
\end{tabular}
\end{table}
\setlength{\extrarowheight}{0\baselineskip}

\paragraph{Lagrangian Duality}
The Lagrangian route to deriving the dual program starts by forming the Lagrangian function by introducing Lagrangian multipliers. The dual program is then the problem of maximizing the infimum of the Lagrangian over the primal variables subject to the Lagrangian multiplier sign constraints (for minimization primals).
\begin{enumerate}
  \item Take as input the primal program Model \ref{eq:primal}.
  \item Form the Lagrangian by introducing multipliers $y_j$. The example below will use the sign convention $y_j\geq0\;\;\forall j\in\mathcal{I}_\leq$, $\;\;y_j\leq0\;\;\forall j\in\mathcal{I}_\geq$,
  $\;\;y_j\in\mathbb{R}\;\;\forall j\in\mathcal{I}_=$ which corresponds to residual convention $b_j-a_j^\top x$. \footnote{Note that the opposite sign convention can be used if using $a_j^\top x -b_j$ for that residual.}
    \begin{align*}
      L(x,y)
      &= c^\top x \;+\; \sum_{j\in\mathcal{I}_\leq} y_j^\top(b_j-a_j^\top x)
      \;+\; \sum_{j\in\mathcal{I}_\geq} y_j^\top(b_j-a_j^\top x)
      \;+\; \sum_{j\in\mathcal{I}_=} y_j^\top(b_j-a_j^\top x)\\
      &= b^\top y + x^\top(c-A^\top y)
      \end{align*}

  \item Form the dual function by taking the infimum of the Lagrangian over $x$:
    \[
      d(y)
      = \inf_{x\in\mathbb R^n} L(x,y)
      = \begin{cases}
          b^\top y & \text{if } c-A^\top y=0 \\
          -\infty & \text{otherwise}
      \end{cases}
    \]
  \item Maximize the dual function subject to the Lagrangian multiplier constraints:
      \begin{equation*}
      \begin{aligned}
        \max_y\enspace &d(y) \\
        \text{s.t.}\enspace & y_j \geq 0 &\forall j \in\mathcal{I}_\leq \\
        & y_j \leq 0 &\forall j \in\mathcal{I}_\geq \\
        & y_j \in\mathbb{R} &\forall j \in\mathcal{I}_=
      \end{aligned}
      \quad\Longrightarrow\quad
      \begin{aligned}
        \max_{y}\enspace &b^\top y\quad \\
        \text{s.t.}\enspace &A^\top y = c \\
        & y_j \geq 0 &\forall j \in\mathcal{I}_\leq \\
        & y_j \leq 0 &\forall j \in\mathcal{I}_\geq \\
        & y_j \in\mathbb{R} &\forall j \in\mathcal{I}_=
      \end{aligned}
      \end{equation*}
\end{enumerate}

\paragraph{Automatic Dualization}
Several software systems allow for the automatic dualization of convex programs including YALMIP in MATLAB \cite{Lofberg2009,MATLAB:2010} and JuMP in Julia \cite{Lubin2023,Julia-2017}. DualSchool uses
\texttt{Dualization.jl} \cite{dualizationjl} from the JuMP ecosystem which implements a standard form-based approach.

\subsection{CGED Implementation Details}\label{appendix:cged_implementation}
Besides the canonicalization steps described in Section \ref{sec:equivalence}, there are several differences in how CGED is implemented compared to the EOR \cite{zhang2025decision} implementation of NGED:
\begin{enumerate}
    \item Variable nodes have only one feature $c_i$ compared to the $c_i$, $l_i$, and $u_i$ in NGED. This is due to the fact that variable bounds are included in the constraint nodes.
    \item Constraint nodes have only one feature $b_j$ compared to the $l_i$, $u_i$ in NGED since in CGED, constraints are reformulated to $a_j^\top x \geq b_j$ rather than $l_j \leq a_j^\top x \leq u_j$. This allows to consider equivalent $l_j\leq a_j^\top x\leq u_j\iff-u_j\leq-a_j^\top x\leq-l_j$ and $a_j^\top x=b_j\iff-a^\top_j x=-b_j$. Note that the EOR \cite{zhang2025decision} implementation differs from NGED as described in \citet{xing2024towards} in that the EOR version does not normalize constraints with less-than sense to greater-than.
\end{enumerate}

\subsubsection{Variable sign canonicalization example}\label{appendix:dsbc}
The treatment of free and double-sided bounded variables in CGED relies on combining the difference-of-positives trick with the variable permutation invariance of GED. In the case of double-sided bounds, it also relies on the inequality sense normalization.
Consider the example ground-truth and candidate pair in Equation \ref{eq:signelim2}.
\begin{equation}\label{eq:signelim2}
    \begin{aligned}
    \min\quad& x_1 + x_2 \\
    \text{s.t.}\quad& x_1+x_2\geq1 \\    & 1\leq x_1\leq2,\;x_2\geq0
    \end{aligned}
    \quad\iff\quad
    \begin{aligned}
    \min\quad& {-}x_1 + x_2 \\
    \text{s.t.}\quad&    -x_1+x_2\geq1\\
 &-2\leq x_1\leq-1,\;x_2\geq0
    \end{aligned}
\end{equation}

The algorithm begins by moving the double-sided bounds to the constraints, normalizing to $\geq$ sense. 
\begin{equation}\label{eq:signelimfree}
    \begin{aligned}
    \min\quad& x_1 + x_2 \\
    \text{s.t.}\quad& x_1+x_2\geq1 \\    & x_1\geq1 \\& -x_1\geq-2\\ & x_1\in\mathbb{R},\;x_2\geq0
    \end{aligned}
    \quad\iff\quad
    \begin{aligned}
    \min\quad& {-}x_1 + x_2 \\
    \text{s.t.}\quad&    -x_1+x_2\geq1\\
 &x_1\geq-2\\&-x_1\geq 1\\&x_1\in\mathbb{R},\;x_2\geq0
    \end{aligned}
\end{equation}
Then the difference of positives transformation is applied to free variables.
\begin{equation}\label{eq:signelimfree2}
    \begin{aligned}
    \min\quad& x_1^+-x_1^- + x_2 \\
    \text{s.t.}\quad& x_1^+-x_1^-+x_2\geq1 \\
    &-x_1^++x_1^-\geq-2 \\
    & x_1^+-x_1^-\geq1\\
    & x_1^+\geq0,\;x_1^-\geq0,\;x_2\geq0
    \end{aligned}
    \quad\iff\quad
    \begin{aligned}
    \min\quad& {-}x_1^++x_1^- + x_2 \\
    \text{s.t.}\quad&    -x_1^++x_1^-+x_2\geq1\\
 &x_1^+-x_1^-\geq-2\\&-x_1^++x_1^-\geq 1\\&x_1^+\geq0,\;x_1^-\geq0,\;x_2\geq0
    \end{aligned}
\end{equation}
Observe that the formulations are identical when swapping $x_1^+$ for $x_1^-$ and vice-versa, which corresponds to a node order permutation that GED handles naturally.

Indeed, the difference-of-positives transformation can in principle be applied to all variables. Then, all bounds can be treated as constraints, removing the need for the explicit $x\leq u \implies x\geq -u$ canonicalization step. However, applying the transformation to all variables would introduce many extra nodes and edges which can needlessly slow down the graph edit distance computation and extend the optimal edit path length. Thus, CGED uses explicit canonicalization for one-side finite bounded variables (the most common case) to only introduce additional variables when handling free and double-bounded variables.

\subsection{Additional details on experimental setup}\label{appendix:exp_setup_cont} 

\paragraph{Dataset Generation} The {\sc DualSchool} samples come from three sources:

\begin{enumerate}
  \item \textbf{2D LPs:} 36 canonical polytopes, each with three distinct objective vectors, ranging from simple shapes (e.g., unit square, triangle) to more complex ones (e.g., hexagon, irregular pentagon).
  \item \textbf{CO Relaxations:} Seven families of combinatorial optimization instances are generated using \textsc{GeCO} \cite{Charfreitag_GeCO}: maximum independent set, multidimensional knapsack, maximum cut, maximum clique, minimum vertex cover, packing, and production planning.
  \item \textbf{LLM4OPT-Derived LPs:} 
    \begin{itemize}
      \item NLP4LP (\cite{ahmaditeshnizi2024optimus}): use the provided \texttt{gurobipy} code directly.
      \item NL4OPT\cite{pmlr-v220-ramamonjison23a}, Easy LP\cite{huang2024mamo}, ComplexOR\cite{xiao2023chain}: these benchmarks only supply an objective value and prompt. Thus, Llama 3.3 is used, following \cite{ahmaditeshnizi2024optimus}, to generate \texttt{gurobipy} formulations for each sample. These formulations are checked using the objective value and re-tried until there is a match. Any instance for which there is no match within five retries is excluded from {\sc DualSchool}.
    \end{itemize}
\end{enumerate}

\paragraph{LLM Configuration} All models run with temperature 0.0, context window size 8192, repeat penalty 1.1, top $k$ 40, top $p$ $0.9$ and min $p$ 0.05. Inference is performed via Ollama\footnote{\url{https://github.com/ollama/ollama}}.

\paragraph{Compute Resources}
The paper used in total about 1000 GPU-hours to run LLM inference for both the experiments presented and those run during development. The evaluations are run on CPU and are relatively fast, adding up to only about 1 CPU-hour. All experiments were run on a node with two Intel Xeon 6426Y (2.5GHz) CPUs and 8 NVIDIA L40S 48GB GPUs.

\section{Additional Experimental Results}\label{sec:more_exp}

\begin{table}[ht]
  \centering
  \caption{Aggregated accuracy results for the {\sc generation} task (continued)}
  \label{tab:llm-benchmark_additional}
  % \resizebox{\textwidth}{!}{%
    \begin{tabular}{llcc@{\hspace{0.75\tabcolsep}}c@{\hspace{0.75\tabcolsep}}cc@{\hspace{0.75\tabcolsep}}c@{\hspace{0.75\tabcolsep}}cc@{\hspace{0.75\tabcolsep}}c@{\hspace{0.75\tabcolsep}}cc@{\hspace{0.75\tabcolsep}}c@{\hspace{0.75\tabcolsep}}c}
      \toprule
       &  &  & \multicolumn{3}{c}{\textbf{CO small}} & \multicolumn{3}{c}{\textbf{Easy LP}} \\
      \cmidrule(lr){4-6} \cmidrule(lr){7-9}
      \textbf{Model} & \textbf{Prompt} & \textbf{Exec\%} & NGED & OBJ & CGED & NGED & OBJ & CGED \\
      \midrule
      \multirow{2}{*}{Mistral - 7B} & 0-shot & 23.0 & 0 & 0 & 0 & 0 & 0 & 0 \\
      & 1-shot & 21.3 & 0 & 0 & 0 & 0 & 0 & 0 \\
      \midrule
      \multirow{2}{*}{Phi 4 - 14B} & 0-shot & 96.2 & 0 & 0 & 0 & 8.1 & 24.8 & \textbf{7.9} \\
      & 1-shot & 94.0 & 0 & 0 & 0 & 5.5 & 12.9 & \textbf{6.0} \\
      \midrule
      \multirow{2}{*}{Gemma 3 - 12B} & 0-shot & 64.8 & 0 & 0 & 0 & 0 & 0.8 & 0 \\
      & 1-shot & 91.4 & 0 & 0 & 0 & 0 & 0 & 0 \\
      \midrule 
      \multirow{2}{*}{Qwen 2.5 – 7B} & 0-shot & 80.9 & 0 & 0 & 0 & 0 & 1.5 & 0 \\
      & 1-shot & 89.3 & 0 & 0 & 0 & 0 & 0.5 & 0 \\
      \midrule
      \multirow{2}{*}{Qwen 2.5 – 14B} & 0-shot & 84.1 & 0 & 0 & 0 & 0 & 5.6 & 0 \\
      & 1-shot & 96.2 & 0 & 0 & 0 & 0 & 0.4 & 0 \\
      \midrule
      \multirow{2}{*}{Llama 3.1 - 8B} & 0-shot & 92.1 & 0 & 0 & 0 & 0 & 1.0 & 0 \\
      & 1-shot & 89.0 & 0 & 0 & 0 & 0 & 0.6 & 0 \\
      \midrule
      \multirow{2}{*}{Llama 3.3 - 70B} & 0-shot & 71.0 & 0 & 0 & 0 & 3.5 & 10.4 & 3.5 \\
      & 1-shot & 99.8 & 0 & 0 & 0 & 0 & 1.0 & 0 \\
      \midrule
      \bottomrule
    \end{tabular}
  % }
\end{table}

\begin{figure}[ht]
     \centering
     \includegraphics[width=\textwidth]{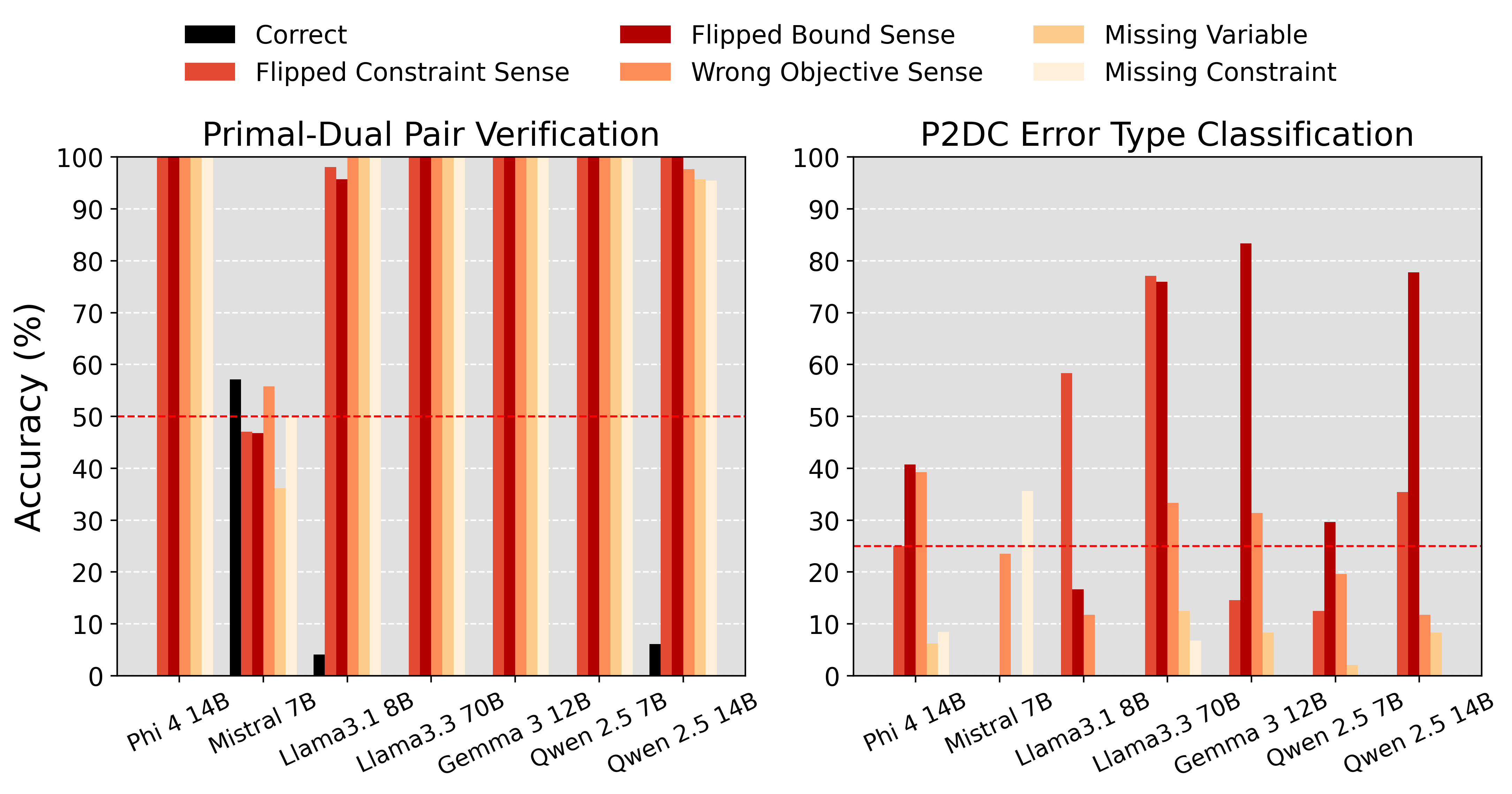}
     \caption{Accuracy for \textsc{Verification} and \textsc{Classification} \textsc{DualSchool} tasks when using structured outputs.}
     \label{fig:verification_acc_structured}
\end{figure}

\subsection{LLMs know the procedure}\label{appendix:theyknow}
To verify that the evaluated models ``know'' how to dualize an LP, the authors prompt and manually evaluate each model's response to ``How do you convert a primal linear program to its dual?'' Since this prompt does not request a highly detailed response nor an example, LLMs are free to remain at a high level where it is easier to describe a valid procedure; details such as converting correctly to its chosen standard form are a single sentence rather than a careful subroutine implementation. In this setting, all models evaluated produced valid procedures except for Gemma 3 - 12B which has an incorrect sign in the objective coefficients and Mistral 7B which forgot to treat primal equality constraints. Notably, all models used a standard form-type method, though with varying standard forms. 

\subsection{Symbolic and Synthetic Verification of LLM P2DC}
Although the main task of {\sc DualSchool} is dualization of specific linear programs, i.e. single instances, one may consider instead prompting the LLMs to generate code that carries out the dualization more generically, i.e. to write Python code implementing symbolic dualization. This section contains experimental results for such a setting, where the LLM is asked to create a function that takes the primal problem data tuple ($A$, $b$, $c$, $l$, $u$, objective sense, constraint senses) and returns its dual as a \texttt{gurobipy} model. Table \ref{tab:llm-benchmark-simple} reports the aggregated accuracy results for this task, for both open and closed models. Similarly to the main {\sc generation} task, the execution accuracy is high -- in fact 100\% -- indicating the LLMs reliably produce executable code. However, the dualization accuracy of the produced routines is very low, with only chatGPT 4o achieving a non-zero CGED.

\begin{table}[ht]
  \centering
  \caption{Aggregated accuracy results for the symbolic P2DC task.}
  \label{tab:llm-benchmark-simple}
  \resizebox{\textwidth}{!}{%
  \begin{tabular}{lcccccccccccc}
    \toprule
    \textbf{Model} & \multicolumn{3}{c}{\textbf{ComplexOR}} & \multicolumn{3}{c}{\textbf{NL4OPT}} & \multicolumn{3}{c}{\textbf{NLP4LP}} & \multicolumn{3}{c}{\textbf{Easy LP}} \\
    \cmidrule(lr){2-4} \cmidrule(lr){5-7} \cmidrule(lr){8-10} \cmidrule(lr){11-13}
     & NGED & OBJ & CGED & NGED & OBJ & CGED & NGED & OBJ & CGED & NGED & OBJ & CGED \\
    \midrule 
    Claude 3.5 & 0 & 0 & 0 & 0 & 0 & 0 & 0 & 0 & 0 & 0 & 0 & 0 \\ 
    Claude 3.7 & 0 & 14.3 & 0 & 0 & 0.5 & 0 & 0 & 3.3 & 0 & 0 & 0 & 0 \\ 
    Gemini 1.5 Flash & 0 & 14.3 & 0 & 0 & 0 & 0 & 0 & 0.4 & 0 & 0 & 0 & 0 \\ 
    Gemini 2.0 Flash & 0 & 14.3 & 0 & 0 & 33.2 & 0 & 0 & 27.4 & 0 & 0 & 0 & 0 \\ 
    Gemini 2.5 Flash & 0 & 0 & 0 & 0 & 0 & 0 & 0 & 0 & 0 & 0 & 0 & 0 \\ 
    Gemma3 12B & 0 & 0 & 0 & 0 & 0 & 0 & 0 & 0 & 0 & 0 & 0 & 0 \\ 
    Gemma3 27B & 0 & 0 & 0 & 0 & 0 & 0 & 0 & 4.1 & 0 & 0 & 0 & 0 \\ 
    Gemma3 4B & 0 & 0 & 0 & 0 & 0 & 0 & 0 & 0 & 0 & 0 & 0 & 0 \\ 
    chatGPT 4o & 7.1 & 7.1 & \textbf{7.1} & 0 & 0.5 & 0 & 0 & 2.9 & 0 & 0 & 0 & 0 \\ 
    chatGPT 4o mini & 0 & 0 & 0 & 0 & 0 & 0 & 0 & 0 & 0 & 0 & 0 & 0 \\ 
    chatGPT 4o mini-high & 0 & 0 & 0 & 0 & 0 & 0 & 0 & 0 & 0 & 0 & 0 & 0 \\ 
    chatGPT O3 & 0 & 0 & 0 & 0 & 0 & 0 & 0 & 0 & 0 & 0 & 0 & 0 \\ 
    Llama3.1 8B & 0 & 7.1 & 0 & 0 & 0 & 0 & 0 & 7.9 & 0 & 0 & 0 & 0 \\ 
    Llama3.2 3B & 0 & 0 & 0 & 0 & 0 & 0 & 0 & 0 & 0 & 0 & 0 & 0 \\ 
    Phi 4 & 0 & 0 & 0 & 0 & 0 & 0 & 0 & 0 & 0 & 0 & 0 & 0 \\ 
    Qwen2.5 14B & 0 & 0 & 0 & 0 & 0 & 0 & 0 & 0 & 0 & 0 & 0 & 0 \\ 
    Qwen2.5 7B & 0 & 0 & 0 & 0 & 0 & 0 & 0 & 0 & 0 & 0 & 0 & 0 \\ 
    \bottomrule
  \end{tabular}}
\end{table}

\subsection{Common mistakes}

An informal analysis reveals the following common mistakes made by LLMs in the \textsc{Generation} task (when the response is almost correct):
\begin{itemize}
    \item \texttt{gurobipy} defaults -- When declaring a new variable in \texttt{gurobipy}, the default lower bound is zero. Sometimes, when attempting to model $x\leq0$, the LLM forgets to set the lower bound to $-\infty$, effectively adding $x=0$ instead.
    \item Incorrect bound senses -- often, the variable bound senses are consistently wrong, i.e. all bounds are flipped compared to ground truth, but the constraint and objective coefficients are not flipped accordingly.
    \item Missing dual variables -- LLMs tend to forget to include dual variables corresponding to primal variable upper bounds when there is a finite lower bound on the same primal variable.
\end{itemize}

\section{Error Types}\label{appendix:P2D_errors} 
 
To systematically inject errors into ground-truth dual programs, the paper considers the following error types.

\paragraph{Wrong Objective Sense} Flip the sense of the dual objective (minimize $\leftrightarrow$ maximize), without adjusting the coefficients.
\begin{equation*}
    \begin{aligned}
    \min\quad& 5x_1 + 4x_2 \\
    \text{s.t.}\quad& 2x_1+3x_2\geq1 \\    & x_1\geq0,\;x_2\geq0
    \end{aligned}
    \quad\implies\quad
    \begin{aligned}
    \color{red}\max\color{black}\quad& 5x_1 + 4x_2 \\
    \text{s.t.}\quad& 2x_1+3x_2\geq1 \\    & x_1\geq0,\;x_2\geq0
    \end{aligned}
\end{equation*}

\paragraph{Missing Variable} Randomly select a variable and delete it from the model.
\begin{equation*}
    \begin{aligned}
    \min\quad& 5x_1 + 4x_2 \\
    \text{s.t.}\quad& 2x_1+3x_2\geq1 \\    & x_1\geq0,\;\color{red}x_2\geq0
    \end{aligned}
    \quad\implies\quad
    \begin{aligned}
    \min\quad& 5x_1 \\
    \text{s.t.}\quad& 2x_1\geq1 \\    & x_1\geq0
    \end{aligned}
\end{equation*}
\paragraph{Missing Constraint} Randomly select a constraint and delete it from the model. 
\begin{equation*}
    \begin{aligned}
    \min\quad& 5x_1 + 4x_2 \\
    \text{s.t.}\quad& \color{red}2x_1+3x_2\geq1\color{black} \\    & x_1\geq0,\;x_2\geq0
    \end{aligned}
    \quad\implies\quad
    \begin{aligned}
    \min\quad& 5x_1 + 4x_2 \\
    \text{s.t.}\quad& x_1\geq0,\;x_2\geq0
    \end{aligned}
\end{equation*}

\paragraph{Flipped Constraint Sense} Randomly select a constraint and change its sense.
\begin{equation*}
    \begin{aligned}
    \min\quad& 5x_1 + 4x_2 \\
    \text{s.t.}\quad& 2x_1+3x_2\geq1 \\    & x_1\geq0,\;x_2\geq0
    \end{aligned}
    \quad\implies\quad
    \begin{aligned}
    \min\quad& 5x_1 + 4x_2 \\
    \text{s.t.}\quad& 2x_1+3x_2\color{red}\leq\color{black}1 \\    & x_1\geq0,\;x_2\geq0
    \end{aligned}
\end{equation*}

\paragraph{Flipped Bound Sense} Randomly select a variable and change the sense of its bound constraint.
\begin{equation*}
    \begin{aligned}
    \min\quad& 5x_1 + 4x_2 \\
    \text{s.t.}\quad& 2x_1+3x_2\geq1 \\    & x_1\geq0,\;x_2\geq0
    \end{aligned}
    \quad\implies\quad
    \begin{aligned}
    \min\quad& 5x_1 + 4x_2 \\
    \text{s.t.}\quad& 2x_1+3x_2\geq1 \\    & x_1\geq0,\;x_2\color{red}\leq\color{black}0
    \end{aligned}
\end{equation*}

\section{Prompt Formats}\label{appendix:prompt} 
\begin{figure}[!h]
    \centering
    \includegraphics[width=\linewidth]{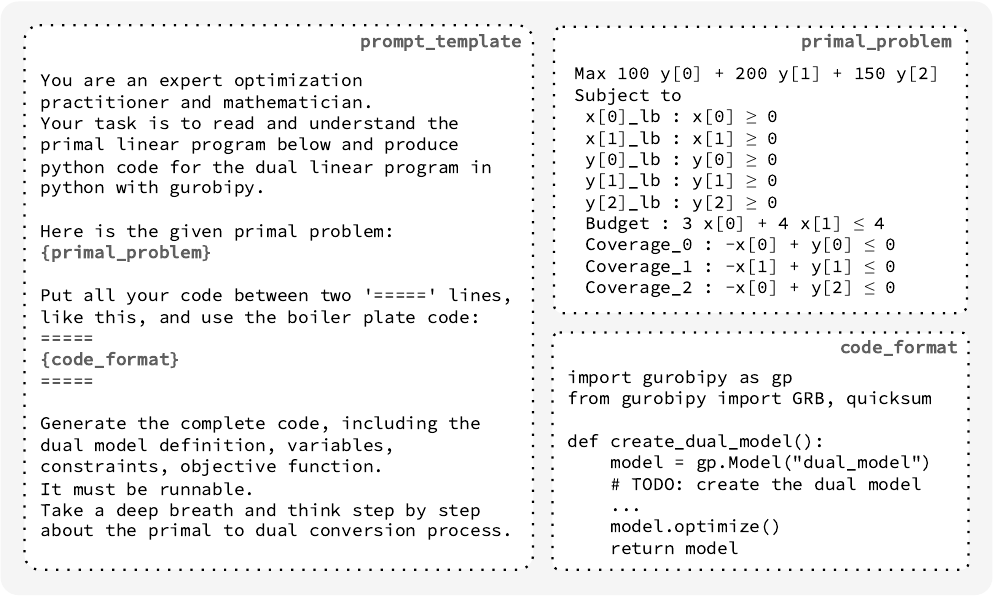}
    \caption{The prompt template used in the Section \ref{sec:experiments} experiments.}
    \label{fig:prompt}
\end{figure}

Figure \ref{fig:prompt} includes a visualization of the prompt format used in the experiments. 

\end{document}